\documentclass[letterpaper, 10pt, journal, twoside]{IEEEtran}

\IEEEoverridecommandlockouts                              

\usepackage[utf8]{inputenc} 
\usepackage[T1]{fontenc}    
\usepackage{hyperref}       %
\usepackage{graphicx}
\usepackage{epsfig}

\usepackage{url}            
\usepackage{booktabs}       
\usepackage{nicefrac}       
\usepackage{microtype}      

\usepackage[dvipsnames]{xcolor}

\usepackage{tikz}
\usepackage{tikzscale} 
\usetikzlibrary{shadows.blur}
\usepackage{tabularx} 
\usepackage{multirow}
\usepackage{colortbl}
\usepackage{amsmath,amssymb,amsfonts}
\usepackage{diagbox} 
\usepackage{caption,subcaption} 
\usepackage{paralist}
\usepackage{glossaries}

\usepackage{pgfplots}
\pgfplotsset{compat=newest}
\usepgfplotslibrary{groupplots}

\usetikzlibrary{external}
\usepackage{dcolumn} 
\newcolumntype{.}{D{.}{.}{-1}} 

\makeglossaries

\newacronym{CoM}{CoM}{Center of Mass}
\newacronym{CoP}{CoP}{Center of Pressure}
\newacronym{RL}{RL}{Reinforcement Learning}
\newacronym{DRL}{DRL}{Deep Reinforcement Learning}
\newacronym{PPO}{PPO}{Proximal Policy Optimization}
\newacronym[longplural={Degrees of Freedom}]{DoF}{DoF}{Degree of Freedom}
\newacronym{PG}{PG}{Policy Gradient}
\newacronym{RBF}{RBF}{Radial Basis Function}
\newacronym{LIP}{LIP}{Linear Inverted Pendulum}
\newacronym{DS}{DS}{Double Support}
\newacronym{SP}{SP}{Support Polygon}
\newacronym{CH}{CH}{Convex Hull}
\newacronym{MDP}{MDP}{Markov Decision Process}
\newcommand\changed[1]{{#1}}
\newcommand\changedFinal[1]{{#1}}

\newcommand{\customtitle}{On the Emergence of Whole-body Strategies\\ from  Humanoid Robot Push-recovery Learning}

\title{\LARGE \bf
\customtitle{}
}

\author{Diego Ferigo$^{*,1,3}$, Raffaello Camoriano$^{*,2}$, Paolo Maria Viceconte$^{1,4}$,\\ Daniele Calandriello$^{2}$, Silvio Traversaro$^{1}$, Lorenzo Rosasco$^{2,5,6}$ and Daniele Pucci$^{1}$
\thanks{Part of this work has been carried out at the Machine Learning Genoa Center, Università di Genova. D. P. acknowledges the support by the An.Dy project, which has received funding from the European Union's Horizon 2020 Research and Innovation Programme under grant agreement No. 731540.
L. R. acknowledges the financial support of the European Research Council (grant SLING 819789), the Center for Brains, Minds and Machines, funded by NSF STC award CCF-1231216, the AFOSR projects FA9550-18-1-7009, FA9550-17-1-0390 and BAA-AFRL-AFOSR-2016-0007 (European Office of Aerospace Research and Development), the EU H2020-MSCA-RISE project NoMADS - DLV-777826, and the NVIDIA Corporation for the donation of the Titan Xp GPUs and the Tesla k40 GPU used for this research.\vspace*{0.2cm}}
\thanks{$^*$Co-first authors $\;\;\;\;\;\;\;\;\;\;\;\;\;\;\;\;\;\;\;\;\;\; $ {\tt\small [name].[surname]@iit.it}}
\thanks{$^{1}$Dynamic Interaction Control, Istituto Italiano di Tecnologia, Genoa, Italy}
\thanks{$^{2}$Laboratory for Computational and Statistical Learning - IIT@MIT, Istituto Italiano di Tecnologia, Genoa, Italy, and
Massachusetts Institute of Technology, Cambridge, MA, USA}
\thanks{$^{3}$Machine Learning and Optimisation, University of Manchester, Manchester, UK}
\thanks{$^{4}$DIAG, Sapienza Università di Roma, Roma, Italy}
\thanks{$^{5}$MaLGa \& DIBRIS, Università degli Studi di Genova, Genova, Italy}
\thanks{$^{6}$Center for Brains, Minds and Machines, MIT, Cambridge, MA, USA\vspace*{0.2cm}}
}

\begin{document}


\maketitle

\begin{abstract}
Balancing and push-recovery are essential capabilities enabling humanoid robots to solve complex locomotion tasks.
In this context, classical control systems tend to be based on simplified physical models and hard-coded strategies. 
Although successful in specific scenarios, this approach requires demanding tuning of parameters and switching logic between specifically-designed controllers for handling more general perturbations.
We apply model-free Deep Reinforcement Learning for training a general and robust humanoid push-recovery policy in a simulation environment.
Our method targets high-dimensional whole-body humanoid control and is validated on the iCub humanoid.
Reward components incorporating expert knowledge on humanoid control enable fast learning of several robust behaviors by the same policy, spanning the entire body.
\changed{We validate our method with extensive quantitative analyses in simulation,} including
out-of-sample tasks which demonstrate policy robustness and generalization, both key requirements towards real-world robot deployment. 
\end{abstract}



\section{Introduction}

Bipeds are those creatures that make use of two legs for moving while maintaining static or dynamic equilibrium.
Balancing is a key prerequisite for any kind of locomotion bipeds may achieve. 
Human evolution determined highly robust  bipedal locomotion, providing enhanced environmental adaptability and fitness with respect to other species. 
Humanoid robots are actuated mechanisms sharing many structural similarities with the human body.
In a world largely crafted by and for humans, they also need to balance for effective operation. 
The challenges posed by bipedal dynamics are manifold. 
Bipeds, compared to other morphologies, are inherently unstable.
Control actions need to account for a narrow support surface and a sparse mass distribution.
Nonetheless, bipedal balancing and locomotion successfully established themselves in nature. 
Therefore, it is reasonable to expect comparable proprioceptive signals to be sufficient for the emergence of similar motor capabilities.

A great variety of methods aiming to solve similar sequential decision-making problems has recently been proposed.
\gls*{DRL} is among the most promising~\cite{lillicrap_continuous_2016}. 
Complex locomotion behaviors can be synthesized by policies trained on sequential interactions with the environment~\cite{heess_emergence_2017}. 
However, this approach poses fundamental challenges when applied to robotics~\cite{dulac-arnold_empirical_2020}.
In particular, collecting the amount of example trajectories required by most state-of-the-art model-free \gls*{DRL} algorithms is unfeasible for current robots~\cite{sigaud_policy_2019}.
A common solution consists in resorting to synthetic data based on rigid-body dynamics, addressing the mismatch introduced by the sim-to-real gap in a subsequent stage~\cite{christiano_transfer_2016,muratore_assessing_2019}.
Nonetheless, learned behaviors often display unnatural characteristics, such as asymmetric gaits, abrupt motions of the body and limbs, or even unrealistic motions exploiting imperfections and glitches in the physical simulator of choice.
These issues significantly limit generalization and transferability to real-world robots.

State-of-the-art methods for bipedal robot control~\cite{feng_optimization_2014} are rooted in control theory and optimal control.
Control architectures are often organized as hierarchies composed of trajectory optimization \changed{\cite{kuindersma_optimization-based_2016}}, simplified model control, and whole-body quadratic programming~\cite{dafarra_torque-controlled_2016, romualdi_benchmarking_2018}. 
While such approaches have achieved considerable results both on simulated and real humanoid robots, they:
\begin{enumerate}[i)]
  \item Rely on an accurate description of the robot dynamics;
  \item Require hand-crafted features for online execution~\cite{dafarra_control_2018};
  \item Present challenges when simultaneously facing different tasks.
\end{enumerate}

\noindent
As concerns push recovery, switching between different strategies (e.g., ankle, hip, stepping, and momentum) is not trivial.

Compared to previous results~\cite{yang_learning_2018}, this work offers the following main contributions: 
\changed{
\begin{itemize}
    \item Demonstration of the emergence of robust momentum-based whole-body push-recovery strategies in addition to ankle, hip, and stepping ones;
    \item Design of reward components to guide learning towards steady-state balancing, with transient push-recovery strategies;
    \item \changedFinal{Definition of a state space -- inspired by floating-base dynamics -- encoding sufficient information for solving the task with no prior knowledge about the desired trajectories.}
\end{itemize}
}


\section{Related Work}
\label{sec:related}

\subsection{Control-theoretic approaches}

Humanoid locomotion control has traditionally been tackled by resorting to simplified models.
In particular, the 3D \gls*{LIP} model is among the most widely employed ones~\cite{kajita_3d_2001}.
Its simplified dynamics proved effective and efficient for trajectory generation in walking, balancing, and push-recovery methods.
%
In the presence of limited perturbations, in-place recovery strategies regulating the \gls*{CoP}~\cite{nori_icub_2015} or the centroidal angular momentum~\cite{traversaro_unied_2017} can be sufficient for recovery.
These include ankle, hip, and foot-tilting strategies 
~\cite{stephens_humanoid_2007,li_humanoid_2017}.
An alternative method, modulating the \gls*{CoM} height was recently proposed \cite{koolen_balance_2016}.
%
Stronger perturbations require the support surface to be enlarged or shifted to ensure that the \gls*{CoP} is kept enclosed in it~\cite{stephens_humanoid_2007}.
A natural way to achieve this is by means of stepping strategies.
To this end, push-recovery stepping controllers based on Zero-Moment Point (ZMP) \cite{vukobratovic_stability_1970} trajectory generation have been proposed~\cite{urata_online_2011}, along with Model Predictive Control (MPC) methods controlling the ZMP while rejecting strong external disturbances~\cite{wieber_trajectory_2006}. Alternatively, footstep planning strategies based on the Capture Point (CP)~\cite{pratt_capture_2006}\changed{, \cite{koolen_capturability-based_2012}} have been employed for position-controlled~\cite{englsberger_bipedal_2011,morisawa_balance_2012} and torque-controlled~\cite{dafarra_torque-controlled_2016} humanoids.
%
Control-theoretic methods significantly improved the state-of-the-art push-recovery performances of humanoids.
Still, they present several limitations:

\begin{enumerate}
    \item Controllers usually encode a single behavior. Being robust to a wide range of perturbations requires complex controller switching;
    \item Robot- and task-specific tuning of the controllers and switching system is a costly trial-and-error procedure;
    \item Simplified models and hard-coded strategies often constrain the attainable behaviors;
    \item MPC-based methods are computationally expensive, hindering real-time deployment.
\end{enumerate}

\subsection{Deep Reinforcement Learning approaches}

In recent years, \gls*{DRL} has been successfully applied to synthesize computationally efficient controllers for complex robotic tasks in a data-driven way, both in simulation and in the real world.
%
%
Quadrupeds have drawn considerable attention in \gls*{DRL} locomotion research, also due to their relatively lower dimensionality and greater stability with respect to bipeds.
Policies trained in simulation have been transferred to real robots via accurate system identification and domain randomization \cite{tan_sim--real_2018,hwangbo_learning_2019}, while the data-efficient Soft Actor-Critic algorithm has been shown to learn robust gait policies from few real-quadruped trials \cite{haarnoja_soft_2018}.
%
%
Remarkably, \gls*{DRL} can also train walking policies for non-humanoid bipedal robots~\changedFinal{\cite{castillo_hybrid_2020}}, including real-world deployment without dynamics randomization \cite{xie_iterative_2019}.

%
%
Other works focus on learning locomotion policies for humanoids.
This setting is more challenging, due to the complex and redundant body structure.
The potential of \gls*{DRL} in this domain was first demonstrated on walking tasks in simulation \cite{schulman_proximal_2017}.
Other methods improve the human-likeness of the behaviors by introducing motion imitation~\cite{peng_deepmimic_2018, bergamin_drecon_2019}.
Still, these methods are more targeted towards benchmarking model-free DRL for continuous control and realistic animation of simplified characters rather than applicability to real humanoid robots.

More recent work has been devoted to training push-recovery \cite{yang_learning_2018} and walking \cite{yang_learning_2020} controllers for accurate humanoid robot models using principles from robot control and transferable observation and reward designs.
The latter approaches, although demonstrating diverse effective behaviors emerging from a single policy, control only the lower body joints. 
\gls*{DRL}-based methods for whole-body humanoid control remain an open problem and have the potential for learning high-dimensional locomotion policies, further improving humanoid capabilities to recover from external perturbations.


\section{Background}
\label{sec:bg}

\subsection{Notation}

\begin{itemize}
    \item $W$ and $B$ denote the world (inertial) frame and the base frame of the robot; $R$ and $L$ denote the frames of the right and left feet.
    \item Given two frames $A$ and $B$, $A[B]$ denotes a new frame with the origin of $A$ and the orientation of $B$.  
    \item $G := G[W]$ denotes the frame with origin on the robot's \gls{CoM} and orientation of the world frame.
    \item $n$ denotes the robot's \glspl{DoF}.
    \item ${}^A \boldsymbol{p}_B \in \mathbb{R}^3$ denotes the coordinates of point $B$ in frame $A$. Superscripts, e.g.\ ${}^A \boldsymbol{p}^{xy}_B$, extract specific coordinates.
    \item Given two frames $A$ and $B$ and a point $C$, the matrix ${}^A R_{B} \in SO(3)$ is such that ${}^A \boldsymbol{p}_C = {}^A R_{B} {}^B \boldsymbol{p}_C + {}^A \boldsymbol{p}_B$.
    \item Given ${}^A R_{B}$, the triplet ${}^A (\psi, \rho, \phi)_B$ denotes the Euler angles of the $z$-$x$-$y$ sequence of intrinsic rotations.
    \item Given $\boldsymbol{w}, \boldsymbol{u} \in \mathbb{R}^3$, we define $\boldsymbol{w}^\wedge = W \in \mathbb{R}^{3 \times 3}$ as the skew-symmetric matrix such that $\boldsymbol{w}^\wedge \boldsymbol{u} = \boldsymbol{w} \times \boldsymbol{u}$, and $W^{\vee} = \boldsymbol{w}$ its inverse.
    \item Given ${}^A \boldsymbol{p}_B$ and three frames $A$, $B$ and $C$, the velocity of the point B w.r.t.\ the origin of frame A, expressed in frame $C$, is ${}^C \boldsymbol{v}_{A,B} = {}^C R_A {}^A \dot{\boldsymbol{p}}_B$.
    \item Given three frames $A$, $B$ and $C$, the angular velocity of frame $B$ w.r.t.\ frame A, expressed in frame $C$ is ${}^C \boldsymbol{\omega}_{A,B} = {}^C R_A ({}^A \dot{R}_B {}^A R_B^\top)^\vee$.
    \item ${}^C \mathbf{v}_{A,B} = ({}^C \boldsymbol{v}_{A,B}, {}^C \boldsymbol{\omega}_{A,B})$ denotes the 6D velocity of frame B w.r.t.\ A expressed in frame C. 
    \item $\boldsymbol{s}, \dot{\boldsymbol{s}} \in \mathbb{R}^n$ denote the joint positions and velocities.
    \item $\mathbf{q} = ({}^W \boldsymbol{p}_B, {}^W R_B, \boldsymbol{s}) \in \mathbb{R}^3 \times SO(3) \times \mathbb{R}^n$ denotes the configuration of the floating-base robot.
    \item $\boldsymbol{\nu} = ({}^B \mathbf{v}_{W,B}, \dot{\boldsymbol{s}}) \in \mathbb{R}^{6+n}$ denotes the system velocity, where the base is represented as \emph{body-fixed} velocity~\cite{traversaro_unied_2017}.
    \item ${}_A \mathbf{f}_F = ({}_A \boldsymbol{f}, {}_A \boldsymbol{m})_F \in \mathbb{R}^6$ denotes the 6D force acting on frame $F$ expressed in frame $A$.
\end{itemize}

In the above definitions, the world frame $W$ is implicitly assumed when $A$ is omitted.

\subsection{\gls*{RL}} 

\changed{We formulate balancing and push recovery 
as a discrete-time \gls*{RL} problem modelled as an infinite \gls*{MDP} with a discounted expected return~\cite{sutton_reinforcement_2018,bertsekas_reinforcement_2019}.}
In this setting, an agent interacts with an environment following a control policy.
At each time step $t$, the agent collects data from the environment in the form of a state $\mathbf{x}_t$. 
The control policy $\pi(\mathbf{a}_t|\mathbf{x}_t)$  selects an action $\mathbf{a}_t$ whose application results in a new state $\mathbf{x}_{t+1}$ and a scalar reward $r_t = r(\mathbf{x}_t,\mathbf{a}_t,\mathbf{x}_{t+1})$ encoding the immediate value of the experienced transition towards solving the target task. 
The interaction generates several trajectories $\tau = \{(\mathbf{x}_0,\mathbf{a}_0,r_0), (\mathbf{x}_1,\mathbf{a}_1,r_1), ...\}$.
The agent's goal is to learn a policy $\pi$ maximizing its expected return 
$
J(\pi) = \mathbb{E}_{\tau \sim \pi}\left [ \sum_{t=0}^{T} \gamma^t r_t\right ]
$
over all possible trajectories $\tau$ induced by the policy, where $T$ is the trajectory length and $\gamma$ the discount factor.

\subsection{\gls*{PG} methods}

A popular class of algorithms addressing expected return maximization for continuous-control tasks
is provided by model-free \gls*{PG} methods \cite{sutton_policy_2000}. 
Given a parameterized policy $\pi_{\boldsymbol{\theta}}(\mathbf{a}_t|\mathbf{x}_t)$, \gls*{PG} methods perform direct gradient-based optimization of $\boldsymbol{\theta}$ over the scalar performance measure :
$$
L^{PG} (\boldsymbol{\theta}) = \hat{\mathbb{E}}_t \left [  \log(\pi_{\boldsymbol{\theta}}(\mathbf{a}_t|\mathbf{x}_t)) \hat{A}_t \right ]
$$
where $\hat{\mathbb{E}}_t$ denotes the empirical mean over a finite batch of trajectories. 
The advantage function $\hat{A}_t = R_t - \hat{V}(\mathbf{x}_t)$ evaluates the advantage of taking action $\mathbf{a}_t$ at state $\mathbf{x}_t$, defined as the difference between the actual return $R_t = \sum_{k=0}^{T-k} \gamma^k r_{t+k}$ collected from $\mathbf{x}_t$ in the sampled trajectory and the current estimate of the value function $\hat{V}(\mathbf{x}_t)$.
Using on-policy samples only, at each iteration of the optimization the gradient of the expected return is estimated by differentiating $L^{PG} (\boldsymbol{\theta})$ and used to update $\boldsymbol{\theta}$. 
Among the available PG algorithms, we employ \gls*{PPO}~\cite{schulman_proximal_2017}, which tackles the instability characterizing the training process in presence of large policy updates by maximizing the objective 
$
L^{CLIP} (\boldsymbol{\theta}) = \hat{\mathbb{E}}_t 
\min \left( \frac{\pi_{\boldsymbol{\theta}}(\mathbf{a}_t,\mathbf{x}_t)}{\pi_{\boldsymbol{\theta}_{old}}(\mathbf{a}_t,\mathbf{x}_t)} \hat{A}_t,\text{clip} \left ( \frac{\pi_{\boldsymbol{\theta}}(\mathbf{a}_t,\mathbf{x}_t)}{\pi_{\boldsymbol{\theta}_{old}}(\mathbf{a}_t,\mathbf{x}_t)}, 1 - \epsilon, 1 + \epsilon \right ) \hat{A}_t
 \right )
$
where $\boldsymbol{\theta}_{old}$ are the pre-update policy parameters and $\epsilon$ the hyperparameter used to clip the policy update.
Maximizing $L^{CLIP} (\boldsymbol{\theta})$ maintains new policies close to old ones while optimizing the objective.


\section{Environment}
\label{sec:env}

\begin{figure}
    \centering
    \includegraphics{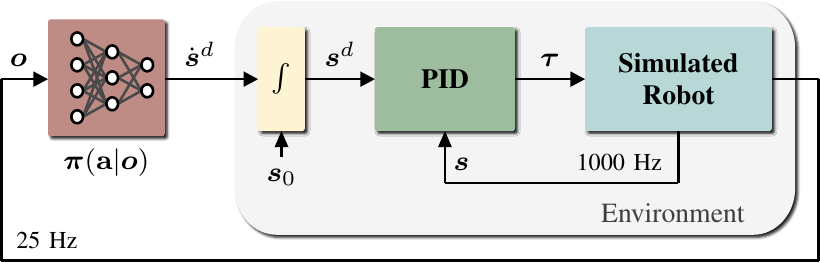}
    \caption{The proposed control system.}
    \label{fig:hierarchical}
    
\end{figure}

The environment is structured as a continuous control task with early termination conditions.
Its dynamics runs in the Ignition Gazebo simulator embedded into the gym-ignition framework~\cite{ferigo_gym-ignition_2020}, compatible with OpenAI Gym~\cite{brockman_openai_2016}. The enabled physics engine is DART~\cite{leeDARTDynamicAnimation2018s}.
We selected iDynTree~\cite{nori_icub_2015} for calculating rigid-body dynamics quantities, using an accurate model of the robot's kinematics and dynamics represented in the following form~\cite{traversaro_unied_2017}:

\begin{equation*}
    M(\mathbf{q}) \dot{\boldsymbol{\nu}} + \boldsymbol{h}(\mathbf{q}, \boldsymbol{\nu}) = B \boldsymbol{\tau} + \sum_{k=1}^{n_c} J_k^\top \mathbf{f}_k
\end{equation*}

where $M(\mathbf{q})$ is the mass matrix, $\boldsymbol{h}(\mathbf{q}, \boldsymbol{\nu})$ the Coriolis and gravity term, $B$ a selector matrix, $\boldsymbol{\tau}$ the joint torques, $n_c$ the number of contacts, $J_k$ and $\mathbf{f}_k$ respectively the Jacobian and the 6D force of the $k$-th contact.

The environment receives actions and provides observations and rewards at 25~Hz. 
The physics and the low-level PIDs run at 1000~Hz.
During training, some properties of the environment are randomized (see Sec.~\ref{sec:env-other}). 

\subsection{Action}

The separation between agent and environment is defined by the  action selection. 
In our nested structure, the policy generates an action $\mathbf{a} \in \mathbb{R}^{23}$ composed of the reference velocities for a large subset of the robot joints (controlled joints), which are then integrated and fed to the corresponding PID position controllers. 
The controlled joints belong to the legs, torso, and arms. Hands, wrists, and neck, which arguably play a minor role in balancing, are locked in their natural positions. 
The policy computes target joint velocities bounded in $[-180,180]$~deg/s at $25$~Hz.
Commanding joint velocities rather than joint positions prevents target joint positions from being too distant from each other in consecutive steps.
Especially at training onset, this would lead to jumpy references that cannot be tracked by the PID controllers, affecting the discovery of the relation between $\mathbf{x}_t$ and $\mathbf{x}_{t+1}$.
The integration process, instead, enables to use a policy that generates discontinuous actions while maintaining continuous PID inputs with no need for additional filters.

\subsection{State}

\begin{table}
    \center
    \caption{Observation components.}
    \label{tab:observation}
    \begin{tabular}{llcc}
        \toprule
        Name & Value & Set & Range \\
        \midrule \rowcolor{black!10}
        Joint positions & $\mathbf{o}_s = \boldsymbol{s}$ & $\mathbb{R}^n$ & $[\boldsymbol{s}_{lb}, \boldsymbol{s}_{ub}]$ \\
        Joint velocities & $\mathbf{o}_{\dot{s}} = \dot{\boldsymbol{s}}$ & $\mathbb{R}^n $ & $[-\pi, \pi]$ \\ \rowcolor{black!10}
        Base height & $\mathbf{o}_h = \boldsymbol{p}_B^z$ & $\mathbb{R}$ & $[0, 0.78]$ \\
        Base orientation & $\mathbf{o}_R = (\rho, \phi)_B$ & $\mathbb{R}^2$ & $[-2\pi, 2\pi]$ \\ \rowcolor{black!10}
        Contact configuration & $\mathbf{o}_c = (c_L, c_R)$ & $\{0, 1\}^2$ & - \\
        \gls{CoP} forces & $ \mathbf{o}_f = (f^{CoP}_L, f^{CoP}_R)$ & $\mathbb{R}^2$ & $[0, mg]$ \\ \rowcolor{black!10}
        Feet positions & $\mathbf{o}_F = ({}^B \boldsymbol{p}_L, {}^B \boldsymbol{p}_R)$ & $\mathbb{R}^6$ & $[0, 0.78]$ \\
        \gls{CoM} velocity & $\mathbf{o}_v = {}^{G}\boldsymbol{v}_{CoM}$ & $\mathbb{R}^3$ & $[0, 3]$ \\
        \bottomrule
    \end{tabular}
\end{table}

The state of the \changed{\gls{MDP}} contains information about the robot's kinematics and dynamics, since no perception is involved. 
It is defined as the tuple $\mathbf{x} := \langle \mathbf{q}, \mathbf{\boldsymbol{\nu}}, \mathbf{f}_L, \mathbf{f}_R \rangle \in \mathcal{X}$. 
The observation, computed from the state $\mathbf{x}$, is defined as the tuple $\boldsymbol{o} := \langle \mathbf{o}_s, \mathbf{o}_{\dot{s}}, \mathbf{o}_h, \mathbf{o}_R, \mathbf{o}_c, \mathbf{o}_f, \mathbf{o}_F, \mathbf{o}_v \rangle \in \mathcal{O}$, where $\mathcal{O} := \mathbb{R}^{62}$.

The observation consists of the following terms:
$\mathbf{o}_s$ are the controlled joints angles in radians, normalized with the hard limits defined in the model description; 
$\mathbf{o}_{\dot{s}}$ are the velocities of the controlled joints, normalized in $[-\pi, \pi]$~rad/s;
$\mathbf{o}_h$ is the height of the base frame, normalized in $[0, 0.78]$~m;
$\mathbf{o}_R$ is a tuple containing the roll and pitch angles of the base frame w.r.t. the world frame, normalized in $[-2\pi, 2\pi]$~rad;
$\mathbf{o}_c$ is a tuple defining whether the feet are in contact with the ground;
$\mathbf{o}_f$ is a tuple containing the vertical forces applied to the local \gls{CoP} of the feet, normalized in $[0, 330]$~N, i.e. the nominal weight force of the robot;
$\mathbf{o}_F$ is a tuple containing the positions of the feet w.r.t. the base frame, normalized in $[0, 0.78]$~m;
$\mathbf{o}_v$ is the linear velocity of the \gls{CoM} expressed in $G$, normalized in $[0, 3]$~m/s.
The exact definition of all the observation terms is reported in Table~\ref{tab:observation}.

Although the agent is trained in simulation, we design it for real-time execution on actual robots.
We carefully select state components that can be either measured or estimated on-board~\cite{nori_icub_2015}. 
To promote policy transfer, we avoid measurements from noisy sensors and values that cannot be estimated with sufficient accuracy. 
In fact, any significant mismatch between simulated and real data would hinder transfer, increasing the reliance on policy robustness. 
We select minimal state components encoding the environment dynamics without affecting learning performance.

\subsection{Reward}

\begin{table*}
    \center
    \caption{Reward function details. Terms with a defined cutoff are processed by the RBF kernel.}
    \label{tab:reward}
    \newcommand{\ck}{\checkmark}
    \setlength\extrarowheight{2pt}
    \begin{tabular}{lcrccrlrr}
        \toprule
        Name & Symbol(s) & Weight & Value $\mathbf{x}$ & Target $\mathbf{x}^*$ & \multicolumn{2}{c}{Cutoff $x_c$}  & SS & DS \\
        \midrule \rowcolor{black!10}
        Joint torques & $r_\tau$ & 5 & $\boldsymbol{\tau}_{step}$ & $\boldsymbol{0}_n$ & 10.0 & Nm & \ck & \ck \\
        Joint velocities & $r_{\dot{s}}$ & 2 & $\boldsymbol{a}$ & $\boldsymbol{0}_n$ & 1.0 & rad/s & \ck & \ck \\ \rowcolor{black!10}
        Postural & $r_{s}$ & 10 & $\boldsymbol{s}$ & $\boldsymbol{s}_0$ & 7.5 & deg & & \ck \\
        CoM $z$ velocity & $r_{v}^z$ & 2 & $\boldsymbol{v}^{xy}_{G}$ & $0$ & 1.0 & m/s & \ck & \ck \\ \rowcolor{black!10}
        CoM $xy$ velocity & $r_{v}^{xy}$ & 2 & $\boldsymbol{v}^{z}_{G}$ & $\omega_0 (\boldsymbol{p}^{xy}_{G} - \boldsymbol{\bar{p}}^{xy}_{hull})$ & 0.5 & m/s & & \ck \\
        Feet contact forces & $\{r_{f}^L, r_{f}^R\}$ & 4 & $\{f^{CoP}_L, f^{CoP}_R\}$ & $m g / 2$ & $m g / 2$ & N & \ck & \ck \\ \rowcolor{black!10}
        Centroidal momentum & $r_h$ & 1 & $\Vert {}_{G} \mathbf{h}_l \Vert^2 + \Vert {}_{G} \mathbf{h}_\omega \Vert^2$ & 0 & 50.0 & kg m$^2$/s & \ck & \ck \\
        Feet CoPs & $\{r_{p}^L, r_{p}^R\}$ & 20 & $\{\boldsymbol{p}_{L, CoP}, \boldsymbol{p}_{R, CoP}\}$ & $\{\bar{\boldsymbol{p}}^{xy}_{L, hull}, \bar{\boldsymbol{p}}^{xy}_{R, hull}\}$ & 0.3 & m & \ck & \ck \\ \rowcolor{black!10}
        Feet orientation & $\{r_{o}^L, r_{o}^R\}$ & 3 & $\{\mathbf{r}^{(z)}_L \cdot \mathbf{e}_z, \mathbf{r}^{(z)}_R \cdot \mathbf{e}_z\}$ & $1$ & 0.01 & - & \ck & \ck \\
        CoM projection & $r_{G}$ & 10 & $\boldsymbol{p}^{xy}_{G}$ & $\in$ CH of support polygon & \multicolumn{1}{r}{-} & - & & \ck \\ \rowcolor{black!10}
        Feet in contact & $r_{c}$ & 2 & $c_L \land c_R$ & 1 & \multicolumn{1}{r}{-} & - & \ck & \ck \\
        Links in contact & $r_l$ & -10 & $c_{l}$ & $0$ & \multicolumn{1}{r}{-} & - & \ck & \ck \\
        \bottomrule
    \end{tabular}
\end{table*}

The reward is a weighted sum of terms that can be categorized as regularizers, steady-state, and transient. \emph{Regularizers} are terms often used in optimal control for the minimization of control action and joint torques. 
\emph{Steady-state} components help to obtain the balancing behavior in the absence of external perturbations, and are active only in \gls{DS}. 
Finally, the \emph{transient} components favor the emergence of push-recovery whole-body strategies. 

The total reward is composed of a weighted sum of scalar components $\sum_i \omega_i r_i$, where $r_i$ is the reward term and $w_i$ its weight.
In order to provide a similar scale for each of them, and therefore improving the interpretability of the total reward, we process the real and vector components with a \gls*{RBF} kernel~\cite{yang_emergence_2017} with a dimension given by a cutoff parameter calculated from the desired sensitivity. 
Appendix~\ref{appendix:kernel} provides a more  detailed description of the kernel. Table~\ref{tab:reward} includes the weights of each reward component and the kernel parameters, if active.

\paragraph*{Regularizers}

\textbf{Joint torques $r_\tau$.}
Torques applied by the PID controllers are penalized. 
\changed{The environment runs at 25~Hz and the low-level controllers at 1000~Hz. Therefore, for each of the 23 joints, 40 torques are actuated between two consecutive environment steps.}
We collect all these torques in a single vector $\boldsymbol{\tau}_{step} \in \mathbb{R}^{23\cdot40}$ and average its elements.
\textbf{Joint velocities $r_{\dot{s}}$.}
Our control scheme  ensures that joint position references are continuous. 
However, PPO explores the action space of joint velocities following the active distributions. 
To promote smoother trajectories, we penalize the norm of the latest action. It can be seen as the minimization of the control effort.

\paragraph*{Steady-state}

\textbf{Postural $r_s$.}
Whole-body humanoid control schemes apply different weights to various control objectives.
The postural is notably one of the most used~\cite{nava_stability_2016}, although it is usually assigned a low priority. 
A postural reward term helps to reach a target posture during balancing instead of relying on local minima found by the learning process. 
This component penalizes the mismatch between the sampled joint configuration and the reference configuration shown in Figure~\ref{fig:icub_q0}.
\textbf{\gls*{CoM} projection $r_{G}$.}
Statically balanced robots, in order to maintain stability, keep the \gls*{CoM} within the \gls*{SP}, defined as the \gls{CH} of their contact points with the ground. 
With the same aim, we introduce a Boolean component rewarding the agent if its \gls*{CoM} ground projection is within the \gls*{SP} induced by the feet. 
For additional safety, we shrink the \gls*{SP} by a 2.5~cm margin all along its perimeter. 
\textbf{Horizontal CoM velocity $r_{v}^{xy}$.}
We define a target horizontal velocity for the \gls*{CoM} as a vector pointing from the \gls*{CoM} projection to the center of the \gls*{SP} $\bar{\boldsymbol{p}}^{xy}_{hull}$.
In order to promote faster motions if the \gls*{CoM} is relatively close to the ground, the magnitude of the target is amplified by a factor $w_0 = \sqrt{g / \boldsymbol{p}^z_{G}}$ derived from the \gls*{LIP} model~\cite{kajita_3d_2001}, where $g$ is the standard gravity.
This component encourages the motion of the \gls*{CoM} projection towards the center of the \gls*{SP}.

\paragraph*{Transient}

\textbf{Feet in contact $r_c$.}
The feet are encouraged to stay on the ground.
In order to promote steps and increase movement freedom, we add a Boolean term marking whether any foot is in contact with the ground.
\textbf{Links in contact $r_l$.}
If any link excluding feet is in contact with the ground, the episode terminates with a negative reward of $-10$ for the terminal state.
\textbf{Whole-body momentum $r_h$.}
Our policy also controls joints belonging to the torso and the arms. 
The momentum generated by the upper body can, therefore, be exploited for balancing and push recovery. 
This term minimizes the sum of the norms of the linear and angular components of the robot's total centroidal momentum ${}_G \mathbf{h}$~\cite{traversaro_unied_2017}.
\textbf{Feet contact forces $r_f$.}
This reward term pushes the transient towards a steady-state pose in which the vertical forces at feet's \glspl*{CoP} $(f^{CoP}_L, f^{CoP}_R)$ assume the value of half of the robot's weight, distributing it equally on the two feet.
\textbf{Feet \glspl*{CoP} $r_p$.}
Beyond the force at the feet \glspl*{CoP}, we also promote their positions to be located at the center of the corresponding sole $\bar{\boldsymbol{p}}^{xy}_{foot, hull}$.
\textbf{Vertical CoM velocity $r_{v}^z$.}
This reward component discourages vertical motion of the \gls*{CoM} of the base link, promoting instead the usage of the horizontal component.
\textbf{Feet orientation $r_o$.}
In early experiments, the policy was converging towards feet tipping behaviors, i.e.\ the feet were not in full contact with the ground. 
Since the terrain is flat by assumption, we discourage tipping by promoting a feet orientation with the soles parallel to the ground. 
If ${}^W R_{foot} = [\mathbf{r}^{(x)}, \mathbf{r}^{(y)}, \mathbf{r}^{(z)}]$ is the rotation between the foot frame and the world, this term promotes the alignment of its third column with the world frame.

\subsection{Other specifications}\label{sec:env-other}

\paragraph*{Initial State Distribution}

The initial state distribution $\rho(\mathbf{x}_0): \mathcal{X} \rightarrow \mathcal{O}$ defines the value of the observation in which the agent begins each episode. 
Sampling the initial state from a distribution with small variance, particularly regarding joint positions and velocities, positively affects exploration without degrading the learning performance. 
At the beginning of each episode, for each joint $j$ we sample its position $s_{j,0}$ from $\mathcal{N}(\mu=s_0, \sigma=10 \, \text{deg})$, where $s_{0}$ represents the fixed initial reference, and its velocity $\dot{s}_{j,0}$ from $\mathcal{N}(\mu=0, \sigma=90 \, \text{deg/s})$.
As a result, the robot may or may not start with the feet in contact with the ground, which encourages the agent to learn how to land and deal with impacts.

\paragraph*{Exploration}

In order to promote exploration beyond the initial state distribution and favor the emergence of push-recovery strategies, we apply external perturbations in the form of a \changed{3D force} to the base frame of the robot.
The applied force vector has a fixed magnitude of 200~N and is applied for 200~ms. 
Considering the weight of the iCub, approximately 33~kg, the normalized impulse sums up to 1.21~Ns/Kg. 
We sample the direction of the applied force from a uniform spherical distribution. 
The frequency of the application is defined as average applications per second, again sampling from a uniform distribution. 
We apply a force on average every 5 simulated seconds.

\paragraph*{Early Termination}

The balancing and push-recovery objectives for a continuous-control task are characterized by an infinite-horizon discounted \changed{\gls{MDP}}. 
During training, however, episodes should stop as soon as the state reaches a subspace from which either it is not possible to recover or it is uninteresting to explore, following an early-termination criterion. 
The state space interesting for our work is where the robot is -- almost -- standing on its feet, therefore we terminate the episodes as soon as it falls to the ground. 
We detect the falling condition when any link but the feet touches the ground plane.

\paragraph*{Domain Randomization}

During the training process, at the beginning of each new episode, the environment performs a domain randomization step. 
The masses of the robot's links are sampled from a normal distribution $\mathcal{N}(\mu=m_0, \sigma=0.2 m_0)$, where $m_0$ is the nominal mass of the link defined in the model description. 
To avoid making assumptions on the material properties of the feet and the ground, we randomize the Coulomb friction $\mu_c$ of the feet by sampling it from $\mathcal{U}(0.5, 3)$.
Finally, since the simulation does not include the real dynamics of the actuators, to increase robustness we apply a delay to the position references that are fed to the PID controllers, sampled from $\mathcal{U}(0, 20)$~ms.


\section{AGENT}

\begin{table}
    \center
    \caption{PPO, policy, and training parameters.}
    \label{tab:training_parameters}
    \newcommand{\ck}{\checkmark}
    \begin{tabular}{c.}
        \toprule
        Parameter & \multicolumn{1}{c}{Value} \\
        \midrule \rowcolor{black!10}
        Discount rate $\gamma$ & 0.95 \\
        Clip parameter $\epsilon$ & 0.3 \\ \rowcolor{black!10}
        Learning rate $\alpha$ & 0.0001 \\ 
        GAE parameter $\lambda$ & 1.0 \\ \rowcolor{black!10}
        Batch size & 10000 \\ 
        Minibatch size & 512 \\ \rowcolor{black!10}
        Number of SGD epochs & 32 \\ 
        Number or parallel workers & 32 \\ \rowcolor{black!10}
        Value function clip parameter & 1000 \\ 
        \bottomrule
    \end{tabular}
\end{table}

\changed{The agent receives the observation $\boldsymbol{o}$ from the environment and returns the action $\mathbf{a}$ defining the reference velocities of the controlled joints. The parameters of the agent are reported in Table~\ref{tab:training_parameters} and further explained below.}

\paragraph*{Learning Algorithm}

We select \gls*{PPO} as candidate learning algorithm, in the variant with both the classic gradient clipping and the minimization of the KL divergence. 

\paragraph*{Policy and Value Function}

The stochastic policy \changed{ $\pi(\mathbf{a}|\boldsymbol{o})$} selects which action to take given a state. 
The value function \changed{$\hat{V}(\boldsymbol{o}_t)$}, instead, estimates the average return when starting from the state $\boldsymbol{o}_t$ and then following the policy for the next steps. 
We represent both the policy and the value function with two different neural networks composed of two fully connected layers, with 512 and 128 units each, followed by a linear output layer. 
The hidden units use a ReLU activation function. 
The networks do not share any layer.


\paragraph*{Distributed Setup}

The chosen \gls*{PPO} algorithm scales gracefully to a setup where the batch samples are collected from multiple workers in parallel.
Our training setup is formed by 32 workers with an independent copy of the environment, and a trainer.
After collecting a batch of 10000 on-policy transitions, we train the neural networks with stochastic gradient descent.
The optimizer uses minibatches containing 512 samples and performs 32 epochs per batch. 
The learning rate is $\lambda = 0.0001$.
Each trial is stopped once it reaches 20~M agent steps, roughly equivalent to 7~days of experience on a real robot.
Worker nodes run only on CPU resources, while the trainer has access to the GPU for accelerating the optimization process.
We use the RLlib~\cite{liang_rllib_2018} framework, OpenAI Gym, and distributed training.


\section{\changed{Results}}
\label{sec:exp}

\subsection{Training performance}

\begin{figure}
    \centering
    \scriptsize
    \includegraphics[width=0.82\columnwidth, height=2.9cm]{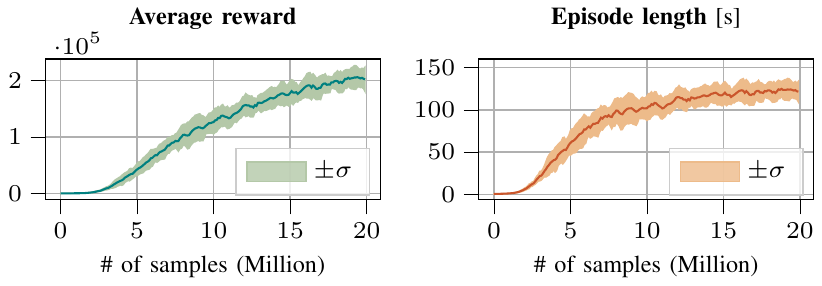}
    \caption{\changedFinal{Learning curves over 11 training runs.}}
    \label{fig:learning_curves}
\end{figure}

Fig.~\ref{fig:learning_curves} reports the learning curves of the average reward and episode duration over 11 independent agent training runs.
Average reward across trials exhibits consistent growth and low variance (Fig.~\ref{fig:learning_curves}, left).
We have also observed increasing values for all individual reward elements during training.
Episode duration improves as well across trials and displays low variance (see Fig.~\ref{fig:learning_curves}, right), approaching maximum episode length more frequently as training progresses.

\subsection{Emerging behaviors}

\begin{figure*}
    \centering
    
    \begin{subfigure}{.15\textwidth}
        \centering
        \includegraphics[height=3.5cm]{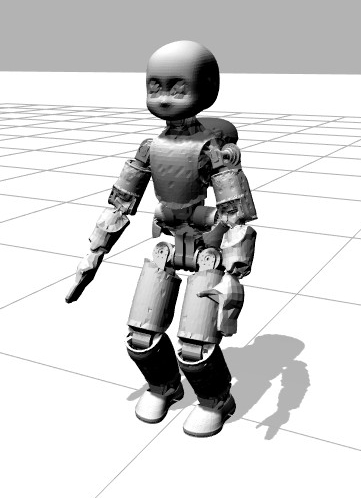}
        \caption{}
        \label{fig:icub_q0}
    \end{subfigure}
    ~
    \begin{subfigure}{.7\textwidth}
        \centering
        \includegraphics[height=3.5cm]{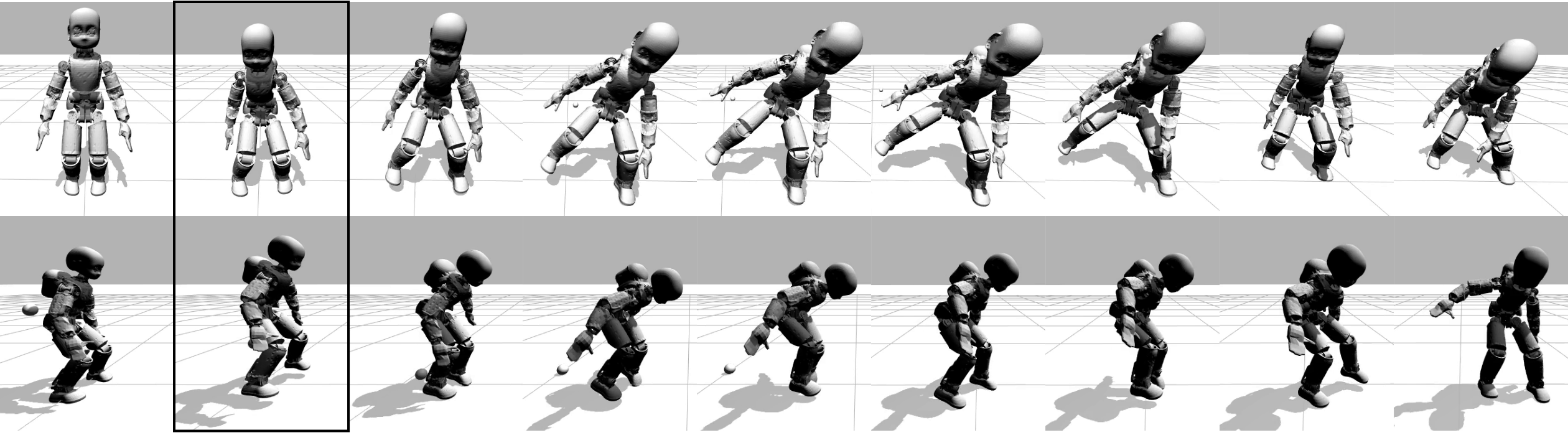}
        \caption{}
        \label{fig:sequences}
    \end{subfigure}
    
    \caption{(a) The initial joint configuration $\boldsymbol{s}_0$. (b) Sequences showing ankle, step, and momentum push-recovery strategies. The robot is pushed by a sphere shot from the left side of the image. Impact takes place in the second frame.}
    \label{fig:q0_and_sequences}
    
\end{figure*}

Controlling the upper body enables rich recovery behaviors that involve the control of the total momentum of the kinematic structure.
We succeed in triggering such behaviors applying external forces during policy training. 
To make force profiles more realistic, instead of applying constant forces for a fixed interval as during training, we throw high-speed objects towards the balanced robot.
Figure~\ref{fig:sequences} shows two characteristic sequences.
\changedFinal{A larger variety of push-recovery strategies are displayed in the supplementary video: \texttt{\url{ https://dic-iit.github.io/emergence-push-recovery-icub/}}}.

\subsection{Deterministic planar forces}

\begin{figure}
    \centering
    \begin{subfigure}{.49\linewidth}
        \centering
        \hspace{-7mm}
        \includegraphics[width=1.1\textwidth]{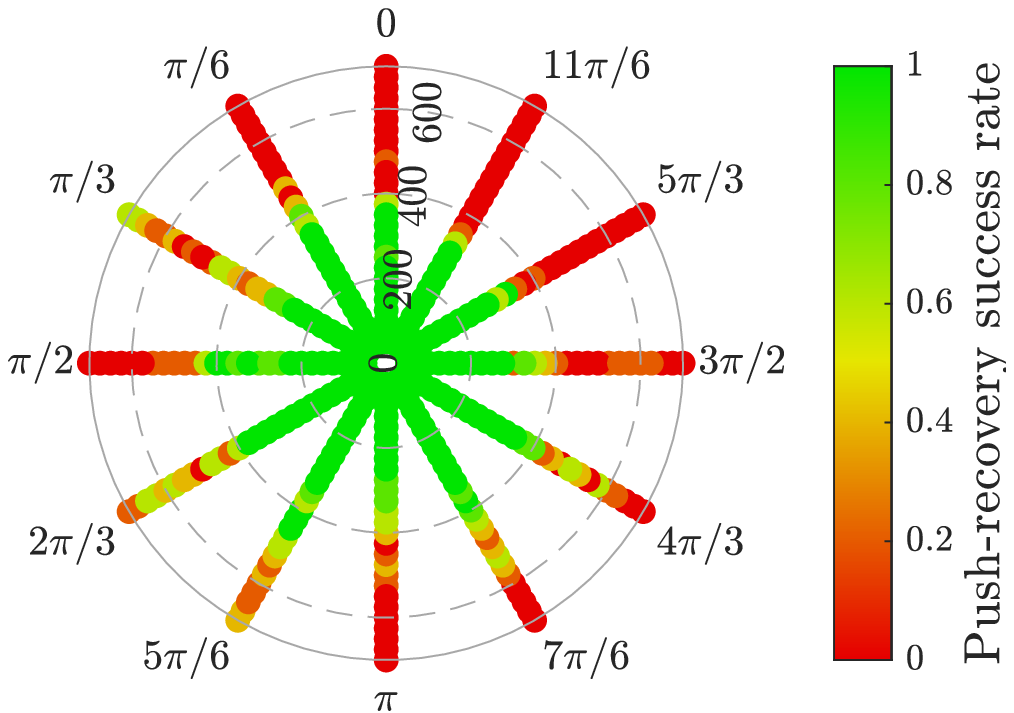}
        \caption{}
        \label{fig:force_polar}
    \end{subfigure}
    \hfill
    \begin{subfigure}{.49\linewidth}
        \centering
        \hspace{-6mm}
        \includegraphics[width=1.1\textwidth]{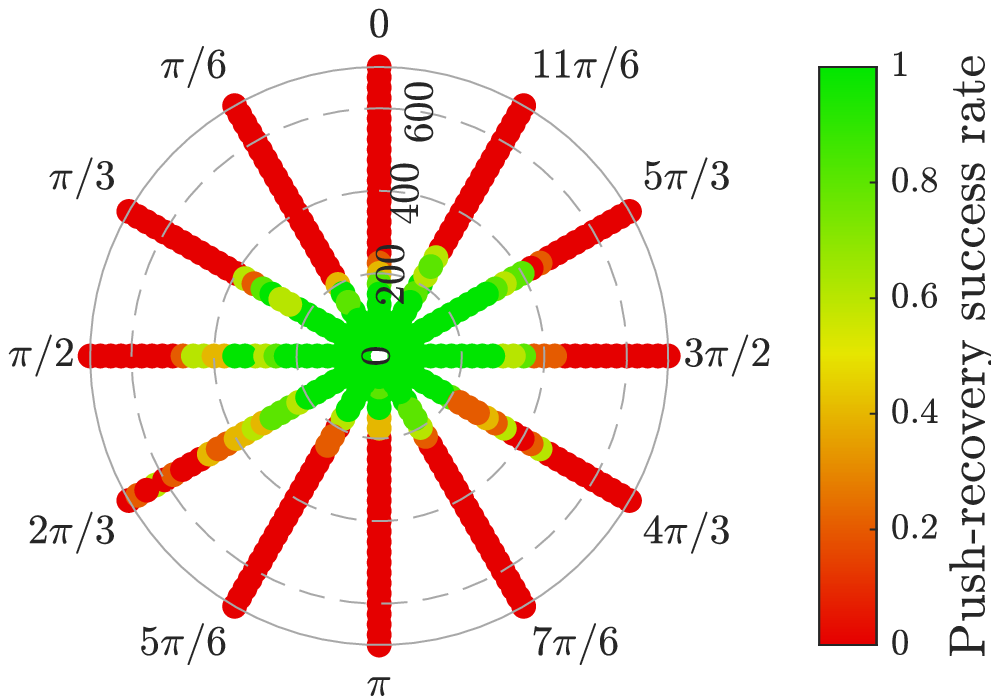}
        \caption{}
        \label{fig:force_polar_low_friction}
    \end{subfigure}
    
    \caption{(a)~Push-recovery success rates on the horizontal plane (forward push: $0$~rad, $\mu_c=1$). (b)~Results with  $\mu_c=0.2$.}
\end{figure}

We evaluate the push-recovery performance from horizontal forces.
Forces are applied for 0.2~s after 3~s from the simulation start, when the robot is stably standing still and front-facing.
Success is defined if the robot is still standing after 7~s.
In Fig.~\ref{fig:force_polar}, success rates for forces pointing in 12 directions are reported.
Magnitudes increase from 50~N to 700~N at 25~N intervals.
5 repetitions are performed for each magnitude and direction, randomizing the initial joints configuration by adding zero-mean Gaussian noise ($\sigma = 2$~deg).
Magnitudes within the training range (0-200~N) are counteracted successfully.
Remarkably, the policy is also robust to out-of-sample forces in all directions in (200-300~N), up to 400~N in some directions.
Moreover, it successfully recovers from pushes in the training range (0-200 N) even with an out-of-sample test friction coefficient $\mu_c=0.2$ (Fig.~\ref{fig:force_polar_low_friction}).

\subsection{Random spherical forces on the base links}

We evaluate policy robustness in challenging scenarios involving sequences of random forces with different combinations of magnitude and duration.
Forces are applied to the base in a random direction more frequently than during training, on average every 3~s.
For each combination, 50 reproducible episodes with different seed initialization and no domain randomization are executed.
Episodes terminate if the robot falls or after 60~s, averaging 20 applications in a full episode.
Our evaluation metric is the number of consecutive forces endured by the robot.
Fig.~\ref{fig:random_forces} reports aggregate results for each combination of magnitude and duration.
No matter their magnitude, forces lasting 0.1~s are properly balanced.
As expected, performances decrease with growing magnitude and duration.
Nevertheless, the agent is able to withstand repeated applications of out-of-sample forces.
For instance, on average it withstands 9 consecutive 300~N 0.2~s applications.

\subsection{Random spherical forces on the chest and elbow links}

\begin{figure}
    \centering
    \scriptsize
    \includegraphics[height=2.7cm, width=\columnwidth]{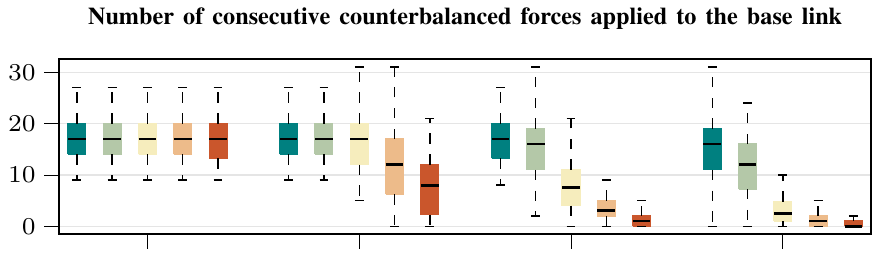}
    \includegraphics[height=2.6cm, width=\columnwidth]{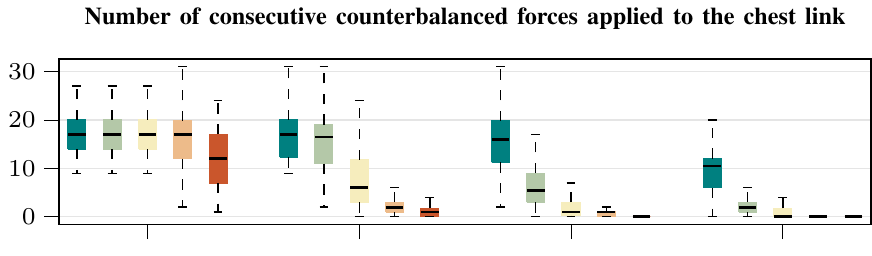}
    \includegraphics[height=3.4cm, width=\columnwidth]{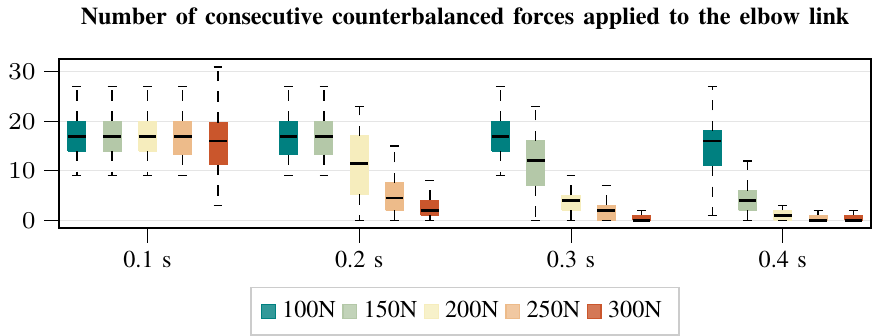}
    \caption{\changedFinal{Consecutive counterbalanced forces in random directions over 50 trials for each combination of magnitude and duration. Forces are applied to the base, chest, and elbow links for an increasing duration.}}
    \label{fig:random_forces}
\end{figure}

We also evaluate robustness of the learned policy to previously unseen forces applied to other links.
Fig.~\ref{fig:random_forces} shows the results obtained on the chest and elbow links.
As expected, forces applied on links which are far from the \gls*{CoM} turn out to be more challenging.
Nevertheless, the policy is able to withstand a good number of them and generalize with good performances.
For instance, it is on average able to recover from 10 consecutive 200~N 0.2~s forces on the elbow link, as opposed to an average of 17 for the base link.
The average number of consecutive counterbalanced forces with the same magnitude and duration decreases to 5 for the chest link.
Notice that the randomness of the interval between two subsequent forces applications leads sometimes to very challenging scenarios in which multiple forces are applied in a very short time span.

\section{Discussion}

\paragraph*{Learning efficiency} 
The overall experience for a single policy training lasts approximately 7~simulated days.
As for other continuous control tasks, model-free \gls*{PG} methods lack sample efficiency. 
There is plenty of room for robot learning research to bridge this efficiency gap.
Indeed, floating-base robots such as iCub can be modeled quite accurately with rigid body dynamics.
Most robots used in research are provided with a dynamic model accurate enough to be exploited as a powerful prior. 
The community has recently proposed interesting model-based algorithms~\cite{
moerland_model-based_2020}
with the potential to improve efficiency and leverage decades of robotics research.

\paragraph*{Low-level control}
Low-level position control is widely adopted in other similar works.
PID controllers have the advantage of being independent of each other and requiring single-joint signals. 
However, besides being difficult to tune, they trade off tracking accuracy with compliance. 
A stiff robot, in the presence of high perturbations, is less robust because even if the planner is whole-body, low-level control is not. 
Whole-body and intrinsically more compliant low-level controllers could be beneficial, although they often operate on the entire underactuated floating-base system.
Properly handling the base references from the policy point of view is yet an uncharted domain.

\paragraph*{Natural behaviour and sim-to-real} 
The emerged push-recovery strategies are not as natural as human ones. 
The policy tends to promote small jumps to full steps, probably due to two factors: the stiffness given by the low-level PIDs, and the difficulty of accurate contact modeling.
As concerns low-level control, actuator dynamics plays a vital role.
Our simulations introduce variable delay but do not saturate joint torques.
Their minimization in the reward does not prevent occasional high torque spikes synthesized by the PIDs.
The integration of more realistic actuator models will be explored in future work.
Regarding contacts, modeling differences between physics engines notably make policies hardly transferable to different engines or the real world. 
The simulator we adopt, Ignition Gazebo, will soon provide a transparent physics engine switch, enabling randomization of the entire engine beyond the common physics parameters.

\section{Conclusions}
\label{sec:conclusions}

We present a \gls*{DRL}-based control architecture capable of learning whole-body balancing and push recovery for
simulated humanoids.
We promote exploration by applying random forces to the kinematic structure, leading to the emergence of a variety of push-recovery behaviors.
Compared to previous works, our policy controls most of the robot's joints, and we show that this contributes to extending the space of recovery \changed{motions to whole-body strategies.
We have shown the results of our architecture controlling 23 \gls*{DoF} of the iCub robot, and showing that our policy can withstand repeated applications of strong external pushes.}

\changed{
Our approach shows different types of limitations.
The PID controllers, while providing a simple low-level control, introduce a stiffness that can prevent natural motion and introduce a joint dynamics that differs from the real platform.
The learning efficiency of model-free algorithms is pretty low and requires days of simulations for a complex behaviour to emerge.
Finally, relying only on state space exploration for finding the expected behaviours requires a carefully designed reward function, that might require a significant effort.
These limitations could be mitigated by introducing prior knowledge in the training scheme, like the usage of model-based whole-body controllers for the low-level and more accurate actuator modeling in simulation, and model-based reinforcement learning.
We plan to explore some of these directions in future work with the aim to bring our policies to the real robot.
}

\appendix[RBF Reward Kernel]
\label{appendix:kernel}

Radial basis function (RBF) kernels are widely employed functions in machine learning, defined as
$$
K(\mathbf{x}, \mathbf{x}^*) = exp \left( -\tilde\gamma ||\mathbf{x} - \mathbf{x}^*||^2  \right) \quad \in [0, 1] ,
$$
where $\tilde\gamma$ is the kernel bandwidth hyperparameter.
The RBF kernel measures similarities between input vectors.
This can be useful for defining scaled reward components.
In particular, if $\mathbf{x}$ is the current measurement and $\mathbf{x}^*$ is the target, the kernel provides a normalized estimate of their similarity.
$\tilde\gamma$ can be used to tune the bandwidth of the kernel, i.e.\ its sensitivity. 
In particular, we use $\tilde\gamma$ to select the threshold from which the kernel tails begin to grow.
Introducing the pair $(x_c, \epsilon)$, with $x_c, \epsilon \in \mathbb{R}^+$ and $|\epsilon| \ll 1$, we can parameterize $\tilde\gamma = -ln(\epsilon) / x_c^2$.
This formulation results in the following properties:

\begin{enumerate}
    \item $K(\mathbf{x}^*, \mathbf{x}^*) = 1$, i.e.\ when the measurement reaches the target, the kernel outputs 1;
    \item Given a measurement $\mathbf{x}_m$ such that $||\mathbf{x}_m - \mathbf{x}^*|| = x_c$, the kernel outputs $K(\mathbf{x}_m, \mathbf{x}^*) = \epsilon$.
\end{enumerate}

\noindent
In practice, $\epsilon$ can be kept constant for each reward component.
The sensitivity of individual components are tuned by adjusting $x_c$.
We refer to $x_c$ as \emph{cutoff} value of the kernel, since each norm of the distance in the input space bigger than $x_c$ yields output values smaller than $\epsilon$.
This formulation eases the composition of the total reward $r_t$ when reward components are calculated from measurements of different dimensionalities and scales.
In fact, once the sensitivities have been properly tuned for each component, they can simply be weighted differently as 
$
r_t = \sum_i w_i K(\mathbf{x}^{(i)}_t, \mathbf{x}^*) \in \mathbb{R}
$
where $\mathbf{x}^{(i)}_t$ is the $i$-th measurement sampled at time $t$, and $w_i \in \mathbb{R}$ the weight corresponding to the $i$-th reward component.

\bibliographystyle{IEEEtran}
\bibliography{IEEEabrv,zotero}

\end{document}